\documentclass[11pt,a4paper]{article}
\usepackage{arabtex}
\usepackage[hyperref]{ranlp2021}
\usepackage{utf8}
\usepackage{graphicx}
\usepackage[font=small,skip=1.5pt]{caption}
\usepackage{amsmath,amssymb}
\usepackage[utf8]{inputenc}
\setcode{utf8}
\usepackage{times}
\usepackage{latexsym}
\usepackage{arydshln}

\usepackage[titletoc,toc,title]{appendix}
\usepackage{lipsum}
\usepackage{cuted}
\usepackage{microtype}

\aclfinalcopy 

\title{Sentiment-Aware Measure (SAM) for Evaluating Sentiment Transfer by Machine Translation Systems}

\author{
  Hadeel Saadany  \\ 
  RGCL \\
  University of Wolverhampton\\
  Wolverhampton, UK\\
\texttt{h.s.saadany@wlv.ac.uk}
  \\\And
  Constantin Or\u{a}san\\
  
  Centre for Translation Studies\\
  University of Surrey\\
  Guildford, UK\\
  \texttt{c.orasan@surrey.ac.uk}
  \\\AND 
  Emad Mohamed\\  
  RGCL\\
  University of Wolverhampton\\
  Wolverhampton, UK \\
  \texttt{e.mohamed2@wlv.ac.uk}
  \\\And
  Ashraf Tantawy\\
 School of Computer Science and Informatics\\
 De Montfort University\\
  Leicester, UK \\
 \texttt{ashraf.tantavy@dmu.ac.uk}
}

\date{}

\begin{document}
\maketitle
\begin{abstract}
In translating text where sentiment is the main message, human translators give particular attention to sentiment-carrying words. The reason is that an incorrect translation of such words would miss the fundamental aspect of the source text, i.e. the author's sentiment. In the online world, MT systems are extensively used to translate User-Generated Content (UGC) such as reviews, tweets, and social media posts, where the main message is often the author's positive or negative attitude towards the topic of the text. It is important in such scenarios to accurately measure how far an MT system can be a reliable real-life utility in transferring the correct affect message. This paper tackles an under-recognised problem in the field of machine translation evaluation which is judging to what extent automatic metrics concur with the gold standard of human evaluation for a correct translation of sentiment. We evaluate the efficacy of conventional quality metrics in spotting a mistranslation of sentiment, especially when it is the sole error in the MT output.  We propose a numerical `sentiment-closeness' measure appropriate for assessing the accuracy of a translated affect message in UGC text by an MT system. We will show that incorporating this sentiment-aware measure can significantly enhance the correlation of some available quality metrics with the human judgement of an accurate translation of sentiment.


\end{abstract}



\section{Introduction}

Standard quality measures for assessing the performance of machine translation systems, such as BLEU \citep{papineni2002bleu}, are domain agnostic; they evaluate the translation accuracy  regardless of the semantic domain  or linguistic peculiarities of the source text. Consequently, they give equal penalty weight to inaccurate translation of n-grams, which may lead to performance overestimation (or underestimation). For example, the Arabic high-rated Goodreads book review showing the reviewers overall satisfaction of a novel: `\<الرواية رهيبة عيبها الوحيد الجزء الاخير>    (`\textit{The novel is great, its only flaw is the last part}') is mistranslated by the available online translation tool which wrongly outputs a negative sentiment (`\textit{The novel is terrible, its only flaw is the last part}'). Despite the distortion of the affect message, this translation receives an equally high score as a correct translation (\textit{The story is great, its only flaw is the last part}), but which uses \textit{story} instead of \textit{novel}. This is because BLEU mildly penalises the wrong translation as it swaps only one uni-gram (`great') with its opposite (`terrible'). Yet, the mistranslation of this particular uni-gram is a critical error as it is most pivotal in transferring the sentiment, and hence the MT performance is over-estimated.




There have been numerous efforts to address the common pitfalls of n-gram-based evaluation metrics by incorporating semantic and contextual features. However, despite research evidence of  its analytical limitations, BLEU is still the \textit{de facto} standard for MT performance evaluation \citep{tangledbleu,structuredrviewbleu}. Moreover, although the introduction of more semantically-oriented metrics showed a better correlation with human judgement, still the estimation of sentiment preservation in UGC has not yet been investigated. This is partially due to the fact that the domain and linguistic style of the WMTs datasets typically used for metric evaluation (e.g. \textit{newstest2020}, Canadian Parliament, Wikipedia, UN corpus) is quite different than the non-standard noisy UGC where sentiment is the main content of a telegraphic message \citep{wmt18, wmt19, wmt20}. Assessed over these WMT datasets, some metrics manifest an almost perfect correlation with human evaluation on the segment-level  (e.g. WMT20 participating metrics record results of up to 0.97 Pearson correlation on the \textit{newstest2020}) \cite{ wmt20}. However, research has also shown that metrics usually report weaker correlation with low human assessment score ranges \cite{takahashi-etal-2020-automatic, takahashi-etal-2020-grammatical}. 

In this paper, we argue that the high correlation of some metrics may not be replicated with a different domain such as sentiment-oriented UGC, specifically when there is a mistranslation of sentiment-critical word(s). For this reason, we propose a `sentiment-closeness' measure that can accommodate for a better evaluation of the MT system's ability to capture the correct sentiment of the source text.




To present our sentiment quality measure, we first briefly explain in Section \ref{sec:Available Metrics} why metrics commonly used for MT quality estimation may not always be efficient in assessing sentiment-critical translation errors. In Section \ref{sec:Sentiment Evaluation Experiment}, we present an experiment to quantitatively assess the divergence of the analysed metrics from a human judgement of a correct/incorrect translation of an affect message. Section \ref{sec:sam} presents our proposed solution for fine-tuning quality metrics for a better correlation with human judgement. Section \ref{sec:assessing} presents the results of incorporating our sentiment-measure in different quality metrics. In Section \ref{sec:error} we conduct an error analysis of our empirical approach and discuss its limitations. Finally, Section \ref{sec:concl} presents a conclusion on our conducted experiments.

\section{Available Metrics}
\label{sec:Available Metrics}

Automatic evaluation metrics usually take the output of an MT system (\textit{hypothesis}) and compare it to one or several translations produced by human translator(s) (\textit{reference}). Based on their matching methods, the most commonly used automatic metrics can be broadly categorised into: surface n-gram matching and embedding matching. Surface n-gram methods work by calculating exact matching, heuristic matching or an edit distance between the aligned n-grams of the reference and hypothesis translation(s). The embedding methods, on the other hand, calculate
a similarity score between learned token representations, such as contextual embedding vectors, with or without the aid of external linguistic
tools. In the following sections, we briefly explain  the methods behind three canonical metrics as representative of each category. We illustrate why the theoretical foundation of each metric may not be optimum for evaluating the translation of sentiment-oriented UGC.

\subsection{Surface N-gram Matching Metrics}
\label{subsec:surface Metrics}

\paragraph{BLEU} The standard metric for assessing empirical improvement of MT systems is BLEU \citep{papineni2002bleu}.  Simply stated, the objective of BLEU is to compare n-grams of the candidate translation with n-grams of the reference translation and count the number of matches; the more the matches, the better the candidate translation. The final score is calculated using a modified n-gram precision multiplied by a brevity penalty so that a good candidate translation would match the reference translation in length, in word choice, and in word order. The disadvantage of the BLEU metric which is relevant to our present study is that it treats all n-grams equally. Due to its restrictive surface n-gram matching, it does not account for the semantic importance of an n-gram in the context of a text. Accordingly, BLEU would incorrectly give a high score to an MT output if it scores exact match with the reference except for one uni-gram, even if this uni-gram completely changes  the sentiment of a text (e.g. `terrible' and `great' as in the Goodreads example above). Online  built-in MT tools have been shown to frequently transfer the exact opposite sentiment word for some dialectical expressions in UGC translated into English \citep{saadany-orasan-2020-great}. Therefore, the BLEU evaluation of an MT performance would be misleadingly over-permissive in such cases where only one or two sentiment-critical words are mistranslated.


\medskip
\noindent \paragraph{METEOR} METEOR \citep{banerjee2005meteor} incorporates semantic information as it evaluates translation by calculating either exact match, stem match, or synonymy match. For synonym matching, it utilises WordNet synsets \citep{wordnet}.  More recent versions (METEOR 1.5 and METEOR++2.0) apply also importance weighting by giving smaller weight to function words \citep{denkowski2014meteor, guo2019meteor++}. The METEOR score ranges from 0 (worst translation) to 1 (best translation). There are two shortcomings to the METEOR metric which do not make it a robust solution for evaluating sentiment transfer. First, the synonym matching is limited to checking whether  the two words belong to the same synset in WordNet. However, WordNet synonymy classification is different than regular thesauruses. For example, `glad' is a synset of `happy' and hence considered a synonym, whereas `cheerful' is not a direct synset to `happy' and hence would be considered a mismatch by METEOR. The following two examples illustrate further the limitations of using WordNet synonymy by METEOR:

\newtheorem{example}{Example}  
\begin{example}\label{ex-m1}
\item
Scores: [\textbf{METEOR: 0.46}]
\begin{itemize}
    \item Hypothesis: ``\textit{The weather is sunny, what a happy day}"
    \item Reference: ``\textit{The sun is shining, what a cheerful day}"
\end{itemize}
\end{example}

\begin{example}\label{ex-m2}
\item
Scores: [\textbf{METEOR: 0.48}]
\begin{itemize}
    \item Hypothesis: ``\textit{I'm not sure why, but I feel so happy today}"
    \item Reference: ``\textit{I don't get it, but I feel so sad today}"
\end{itemize}
\end{example}

\noindent The METEOR scores for Examples \ref{ex-m1} and \ref{ex-m2} clearly diverge from a human evaluation of a good translation. In the first example, although the translation conveys the correct emotion (`happiness'), it receives a similar METEOR score to the hypothesis in the second example which gives the exact opposite emotion of the source (`happiness' instead of `sadness'). The inadequate scoring is a result of the WordNet taxonomy which causes the metric to equally treat both pairs (`sad', `happy') and (`cheerful', `happy') as non-synonym synsets and  hence unmatched chunks.

The second problem with METEOR which may affect its efficacy in evaluating sentiment transfer relates to its weighting schema for function and non-function words. The following example clarifies the gravity of this problem:
\newline
\vspace{10pt}
\begin{example}\label{ex-1a}
\item
Scores: [\textbf{METEOR: 0.92}]
\begin{itemize}
    
    \item Hypothesis: ``\textit{If he had blown himself up in your country, God would forgive him}"
    \item Reference: ``If he had blown himself up in your country, God would \textbf{not} forgive".
\end{itemize}
\end{example}

\noindent The hypothesis in Example \ref{ex-1a} is the translation output of Twitter's built-in MT system for an Arabic tweet commenting on a terrorist attack\footnote{\url{https://twitter.com/gaston810/status/673950532340465664}}. The MT failure to translate the negation marker flips the sentiment of the author from `anger' against the terrorist to `sympathy'. Despite this, the METEOR score is 0.92 which is within the highest upper bound, ranking it as a good translation. On the other hand, the METEOR score for a correct translation of sentiment with a negation marker is 0.93. The main culprit for this inaccurate scoring is the lexical weighting which causes the metric not to penalise the missing of a negation marker which produces a sentiment-critical error. Due to the grave consequences of such mistranslations, it becomes critical to have a sentiment-sensitive metric that is capable of spotting similar errors.


 
\subsection{Embedding-based Metrics}
\label{subsec:embedding Metrics}

\noindent\textbf{BERTScore} \quad Recently embedding-based metrics have proven to achieve the highest performance in recent WMT shared tasks for quality metrics (e.g. \citet{ bleurt, yisi20, 2020mee}). We take BERTScore as a representative metric for this approach \cite{bertscore}.  BERTScore computes a score based on a pair wise cosine similarity between the BERT contextual embeddings of the individual tokens for the hypothesis and the reference \citep{bert}. Accordingly, a BERTScore close to 1 indicates proximity in vector space and hence a good translation. The main problem with embedding-based metrics, such as BERTScore, is that antonyms contain similar distributional information since they usually occur in similar contexts.  Example \ref{ex-1} illustrates this point:


\begin{example}\label{ex-1}
\item
Scores: [\textbf{BERTScore: 0.85}]
\begin{itemize}
    \item Hypothesis: ``\textit{What is this amount of anger, I don’t understand!}"
    \item Reference: ``\textit{What is this amount of happiness, I don’t understand!}"
\end{itemize}

\end{example}

Example \ref{ex-1} shows the mistranslation produced by Twitter's \textit{Translate Tweet} tab of an Arabic tweet. Although the sentiment polarity is flipped in the candidate translation above, the hypothesis receives a BERTScore of 0.85 which indicates a high cosine similarity to the reference in vector space and hence a good translation. Clearly, the metric score is not comparable to a human perception of the emotion reflected by the source. 

\begin{figure}[h]
    \centering
    \includegraphics[scale=0.35,trim={0.1cm 0cm 0cm 0cm},clip]{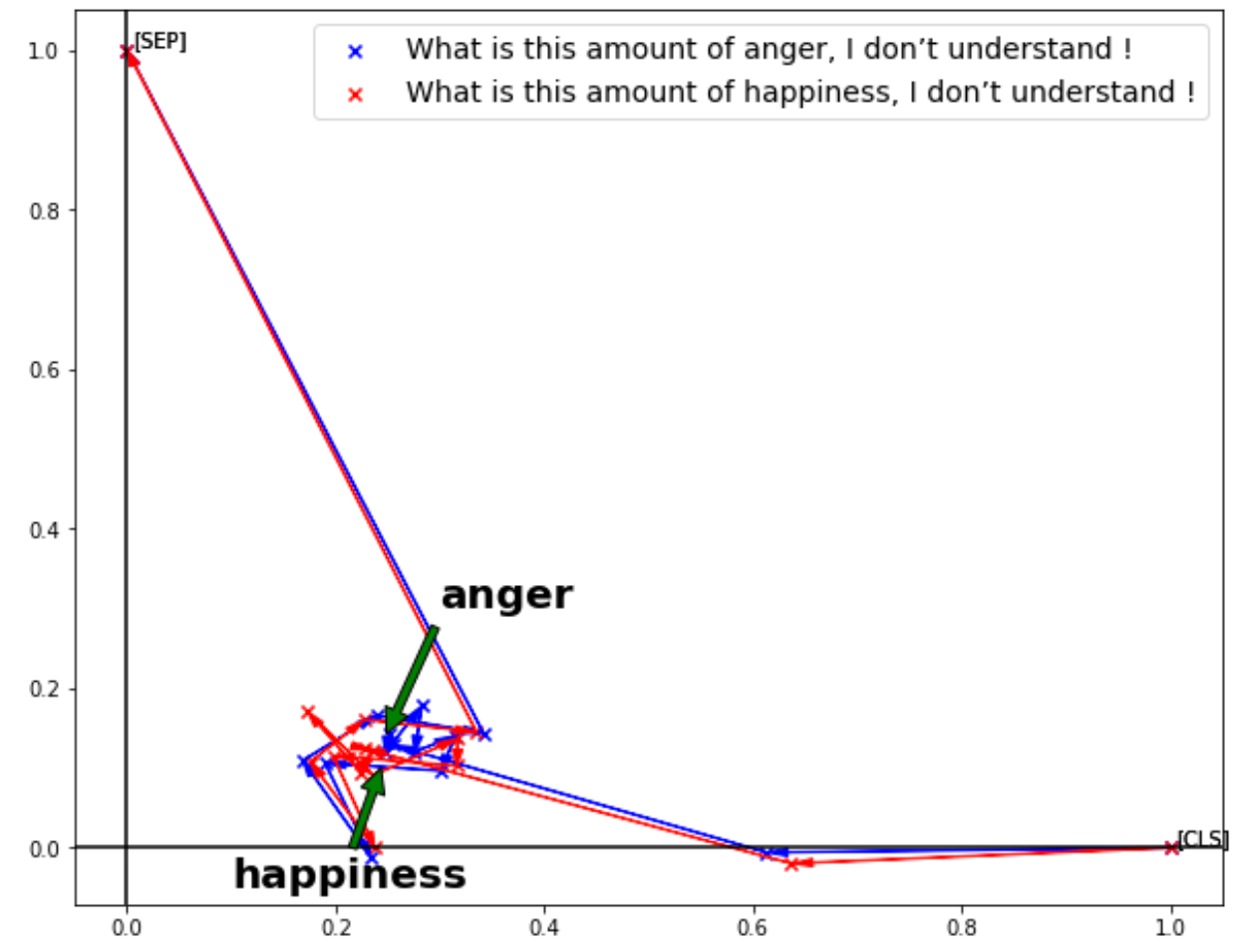}
    \caption{Visualisation of Contextual Embeddings of Sentences in Example \ref{ex-1}}
    \label{bertvis}
\end{figure}

Figure \ref{bertvis} illustrates the reason behind this misleadingly high score.  Figure \ref{bertvis} is a  2-D visualisation of BERT's contextual embedding vectors for the hypothesis translation (in blue) and the reference (in red). Both sentences are very close in the embedding space due to the exact match of their individual tokens. The only mismatch is between the antonyms `happiness' and `anger'. As shown in the figure, the pre-trained embedding vectors of the opposite polarity nouns are also quite close because of their common contextual information. An embedding metric such as BERTScore, therefore, may not penalise antonyms which typically occur in similar contexts. 

Recently, there have been different approaches to overcome some distributional problems of contextual embeddings. \citet{sbert}, for example, introduce SBert, a modification of the pretrained BERT network, which should mitigate the antonymy problem. They use Siamese network structures where the embeddings of similar sentence pairs are independently learned via two parallel transformer architectures. We measured how far this technique could solve the opposite sentiment problem by measuring the SBert sentence similarity of the hypothesis and reference in Example \ref{ex-1}. The cosine similarity of the SBert sentence embedding vectors for the hypothesis and reference in Example \ref{ex-1} reached 0.61. A correct translation of the reference, however, such as \textit{`What is all this cheerfulness, I don't understand'} has a cosine similarity score of 0.79. This small similarity difference (0.18) would be misleading if taken as an evaluation of how far the sentiment poles in the two hypotheses are different. A more sentiment-targeted measure is needed for assessing mistranslations due to this antonymy problem.

In the following section, we conduct an experiment to quantify the divergence of the above mentioned metrics from the human perception of a proper sentiment transfer in a translated text.

\section{Evaluation of Mistranslated Sentiment}
\label{sec:Sentiment Evaluation Experiment}


We have shown in the previous section that three canonical MT evaluation metrics do not give a penalty proportional to sentiment-critical errors on segment-level by an MT online tool. In order to quantify how far the aforementioned quality metrics diverge or correlate with human judgement of sentiment transfer, we measure the performance of each metric on a dataset of tweets that had sentiment-critical translation errors. To compile this data, first we used  the Twitter built-in translation system (Google API) to translate a dataset of tweets annotated for sentiment\footnote{The dataset is collected from SemEval 2018 shared sentiment detection Task  \citep{SemEval2018Task1}. }. The source dataset amounted to $\approx$7,000 tweets in three languages: English, Arabic and Spanish. We translated the Arabic and Spanish into English and the English was translated into Spanish and Arabic.


To extract instances where the MT system failed to translate the sentiment correctly, we built an English sentiment-detection classifier by fine-tuning a Roberta XML model \citep{roberta} on an English dataset of 23,000 tweets annotated for sentiment\footnote{The English dataset is collected from different sentiment detection tasks \cite{2019yisi,wassa}.}. The English classifier was used to predict the sentiment of the Google API output for the translation of the Arabic and Spanish tweets into English and the English back-translation of the English tweets translated into Spanish and Arabic. The classifier’s predicted sentiment was compared to the gold standard emotion of the source text, and instances of discrepancy were extracted as potential mistranslations of sentiment. Finally, from the extracted instances, we manually built a translation quality evaluation dataset. 

The quality evaluation dataset, henceforth QE, consisted of target tweets where the error is exclusively a mistranslation of the sentiment-carrying lexicon. In these tweets, the mistranslations either completely flip the sentiment polarity of the source tweet, similar to Examples \ref{ex-1a} and \ref{ex-1} above, or transfer the same polarity but with a mitigated sentiment tone. The tweets with exclusive sentiment translation errors amounted to 300 tweets. We also added 100 tweet/translation pairs where the MT system transfers the correct sentiment. Reference translations of the QE dataset were created by native speakers of the respective source languages. Essentially, the reference translations aimed at correcting chunks that caused a distortion of the affect message and retained as many of the hypothesis n-grams as possible to detect how far each metric is sensitive to sentiment mismatching and not to the mismatch in other non-sentiment carrying words. The translators were also asked to assign a score to each pair of source-hypothesis tweet, where 1 is the poorest sentiment transfer and 10 is best sentiment transfer. The average scores of annotators were taken as the final human score\footnote{Annotators were computational linguists working on MT research.}.

\begin{figure}[htbp]
    \centering
    \includegraphics[scale=0.65,trim={0.2cm 0cm .2cm 0cm},clip]{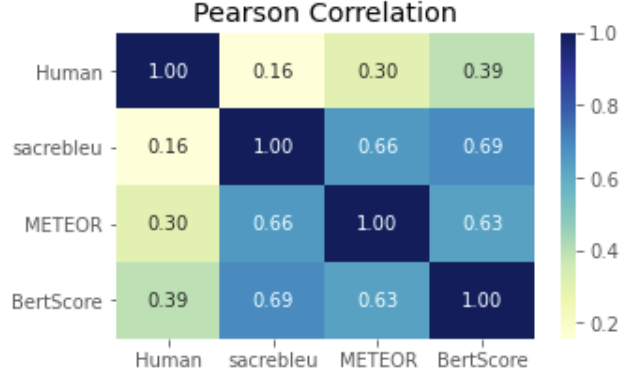}
    \caption{Absolute Pearson correlations with segment-level human judgements for the QE dataset}
    \label{pearson}
\end{figure}

\begin{figure}[htbp]
    \centering
    \includegraphics[scale=0.65,trim={0cm .7cm .3cm 0.20cm},clip]{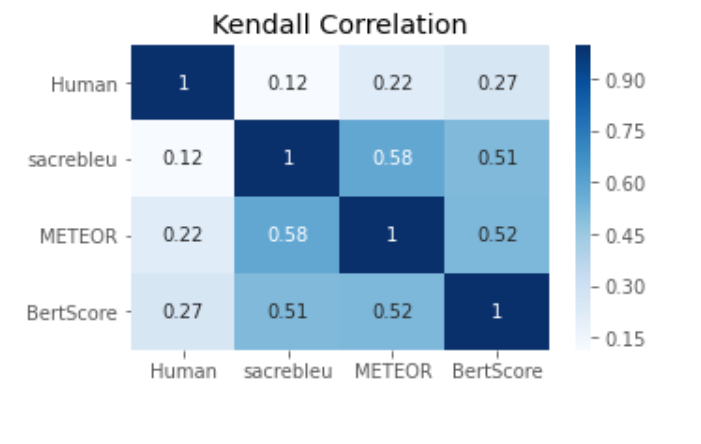}
    \caption{Kendall correlations with segment-level human judgements for the QE dataset}
    \label{kendall}
\end{figure}

To quantify the ability of the three metrics explained in Section \ref{sec:Available Metrics} to assess the transfer of sentiment in the QE dataset, we compared their scores of the translation hypotheses with the human judgement scores for sentiment transfer on the segment-level. We followed the WMT standard methods for evaluating quality metrics and used absolute Pearson correlation coefficient $r$ and the Kendall correlation coefficient $\vert$$\tau$$\vert$ to evaluate each metric's performance against the human judgement. Figures \ref{pearson} and \ref{kendall} show heatmaps visualising the Pearson and Kendall correlation coefficients for the studied metrics and the human scores, respectively.


As seen from Figures \ref{pearson} and \ref{kendall}, both BERTScore and METEOR achieve a better correlation with the human judgement than BLEU\footnote{We use the Sacrebleu implementation of the BLEU score for all the experiments \cite{SACREBLEU}.} which achieves only 0.16 and 0.12 Pearson and Kendall correlations, respectively. However, the relatively overall low correlations (max $r$ = 0.39 and  max $\vert$$\tau$$\vert$  = 0.27)  raise important doubts as to the reliability of these accepted metrics for ranking MT systems which translate sentiments. Bearing in mind that 75\% of the segments in the QE dataset have critical translation errors that seriously distort the sentiment, the low correlation results highlight the need for a sentiment-targeted measure that can improve a metric's efficacy in capturing mistranslated sentiment by an MT system in real-life scenarios.

\section{Sentiment-Aware Measure (SAM) for Machine Translated UGC}
\label{sec:sam}

In this section, we propose a new measure for assessing MT performance that takes into account the sentiment similarity between the MT system translation and the reference. This sentiment measure should be used as a fine-tuning tool to adjust a quality score in cases where it is used to assess the translation quality of sentiment-oriented text. The SAM score is calculated by using the SentiWord dictionary of prior polarities \citep{sentiwords}. SentiWord is a sentiment lexicon that combines the high precision of manual lexica and the high coverage of automatic ones (covering 155,000 words). It is based on assigning a `prior polarity' score for each lemma-POS in both SentiWordNet and a number of human-annotated sentiment lexica \citep{sentiwordnet,warr}. The prior polarity is the out-of-context positive or negative score which a lemma-POS evokes. It is reached via an ensemble learning framework that combines several formulae where each lemma-POS is given the score that receives the highest number of votes from the different formulae. SentiWord prior polarity scores have been proven to achieve the highest correlation with human scores in sentiment analysis regression and classification SemEval tasks \citep{sentiwords}.

We assume that our sentiment adjustment factor, SAM, is proportional to the distance between the sentiment scores of the unmatched words in the system translation and the reference, the higher the distance the greater the SAM adjustment. To calculate the SAM score, we designate the number of remaining mismatched words in the system translation and reference translation by $m$ and $n$, respectively. We calculate the total SentiWord sentiment score for the lemma-POS of the mismatched words in the translation and reference sentences using a weighted average of the sentiment score of each mismatched lemma-POS. The weight of a hypothesis mismatched word $w_h$  and a reference mismatched word $w_r$ is calculated based on the sentiment score of its lemma-POS, $s$, as follows:
\begin{align}
    w^i_h &= |s_i| \qquad i=1,2,\ldots,m. \\
    w^i_r &= |s_i| \qquad i=1,2,\ldots,n.
\end{align}
Then the total sentiment score for hypothesis $S_h$ and reference $S_r$ is given by:
\begin{align}
    S_h &= \sum_{i=1}^m \alpha_i s_i, \quad \alpha_i = \frac{w_h^i}{\sum_{i=1}^{m}w^i_h} \\
    S_r &= \sum_{i=1}^n \beta_i s_i, \quad \beta_i = \frac{w_r^i}{\sum_{i=1}^{n}w^i_r}
\end{align}
The normalised SAM adjustment is given by:
\begin{align}
    p = \frac{|S_r - S_h|}{2}
\end{align}
and the translation final score will be given by:
\begin{align}
    \text{Score} = C_{hr} \left( 1 - p \right)
\end{align}
where $C_{hr}$ can be any metric's matching score between a translation hypothesis and a reference segment. For this experiment, we will measure $C_{hr}$ as the BLEU, METEOR and BERTScore scores. 
To illustrate how SAM score adjusts a metric score with respect to the transfer of sentiment, table \ref{tab1} shows the SAM score adjustment for Examples \ref{ex-1a} and \ref{ex-1} above.
 
\begin{table*}[htp]

\caption{Calculating the SAM adjustment for Examples \ref{ex-1a} and \ref{ex-1}}\label{tab1}
\centering
\begin{tabular}{l|l|l|l|l|l|l|l}
\hline
\textbf{Example} &
  \textbf{$w_h$} &
  \textbf{$w_r$} &
  \textbf{$S_h$} &
  \textbf{$S_r$} &
  \textbf{$p$} & 
  \textbf{$C_{hr}$} &
  \textbf{Score+SAM}\\ \hline
 \textbf{\ref{ex-1a}} &
  him\#a &
  not\#r &
  0 &
  -1.0 &
  0.5 & 
  0.92 &
  0.46 \\ \hdashline[2pt/1pt]
  \textbf{\ref{ex-1}} &
  anger\#n &
  happiness\#n &
  -0.669 &
  0.856 &
  0.762 & 
  0.85 &
  0.20 \\ \hline

\end{tabular}
\end{table*}

As can be seen from Table \ref{tab1}, the metric score $C_{hr}$ (here METEOR and BERTScore, respectively) is significantly reduced due to the sentiment distance between the mismatched words ($w_h$, $w_r$) as well as their sentiment weights ($S_h$, $S_r$). The reduced scores are more representative of the distortion of sentiment produced by the MT system in Examples \ref{ex-1a} and \ref{ex-1}. The SAM adjustment, therefore, is a targeted-measure that can fine-tune a metric according to the `sentiment-closeness' of the translation and the reference\footnote{More examples of SAM adjustments are in Appendix \ref{appendixa}}.

\section{Assessing SAM performance} \label{sec:assessing}

\begin{figure}[h!]
    \centering
    \includegraphics[scale=0.55,trim={0.2cm 0.20cm .3cm 0cm},clip]{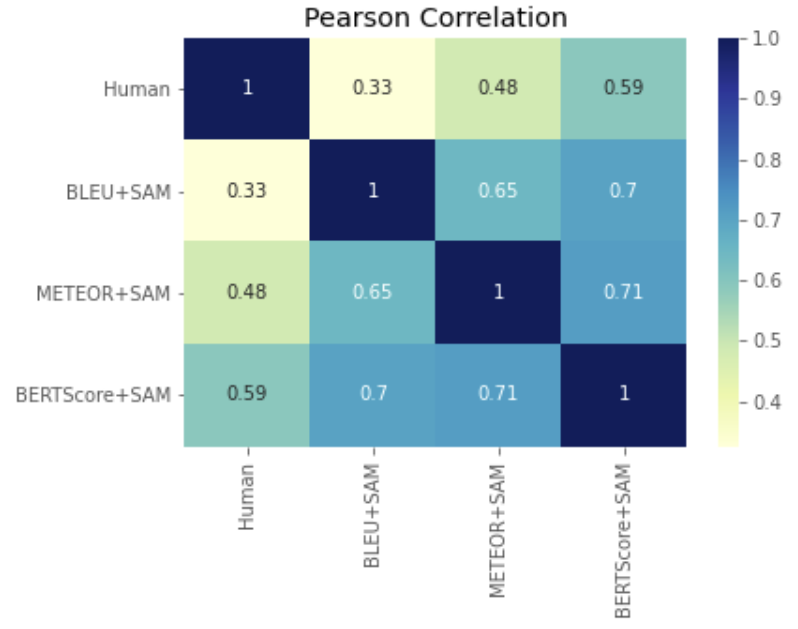}
    \caption{Absolute Pearson correlations with SAM Adjustment for the QE dataset}
    \label{pearsonwith}
\end{figure}

\begin{figure}[h!]
    \centering
    \includegraphics[scale=0.55,trim={0.5cm 0.38cm .2cm .1cm},clip]{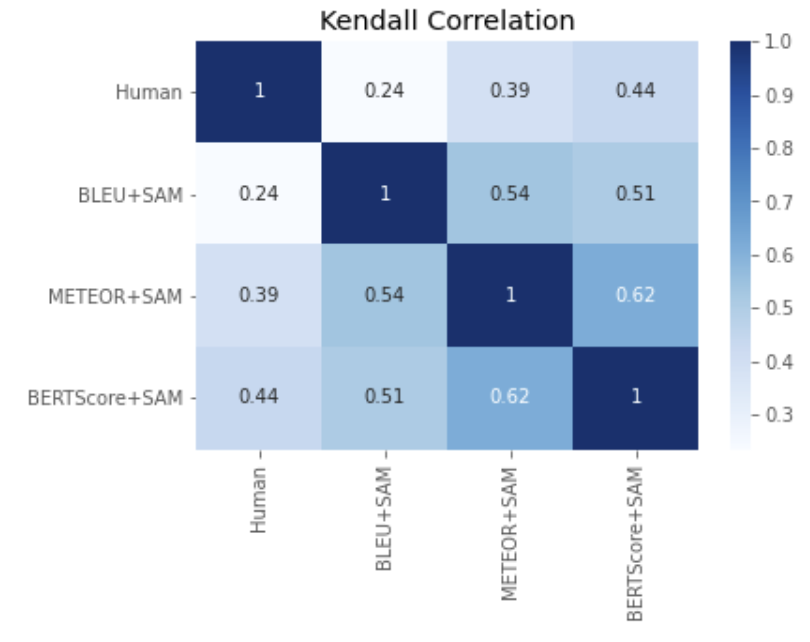}
    \caption{Kendall correlations with SAM adjustment for the QE dataset}
    \label{kendallwith}
\end{figure}

To check how far the SAM measure can improve the evaluation of sentiment, we assessed the performance of the three chosen metrics  on the QE dataset utilised in the experiment in \mbox{Section \ref{sec:Sentiment Evaluation Experiment}} with the SAM adjustment. Figures \ref{pearsonwith} and  \ref{kendallwith} show heatmaps of the Pearson and Kendall correlations of the SAM adjusted metrics with the human judgement in the QE dataset.


\vspace{-.7pt}

Compared to metric scores without the SAM adjustment, Figures \ref{pearsonwith} and \ref{kendallwith} show that the combination of SAM with the three metrics consistently leads to an improvement in each metric's correlation with the human judgement of a correct sentiment transfer in our dataset. Overall, BERTScore records the highest Pearson correlation coefficients and Kendall rank dependence (0.59 and 0.44, respectively). This means that it is better able to penalise critical translation errors in our QE dataset. Moreover, compared to their scores without SAM, both BERTScore and METEOR record 20\% and 18\%  higher correlation, respectively. It is also worthy of notice, that although the Pearson correlation of BLEU improves from 0.16 to 0.33 with the SAM adjustment, the correlation is still relatively small. Knowing that BLEU score is usually the gate-keeping tool for accepting improvement in MT research, the results cast doubt on the efficacy of this non-semantic method to penalise sentiment-critical translation errors.


\section{Error Analysis and Limitations}
\label{sec:error}

Human sentiment can be expressed in intricately subtle ways so that the mistranslation of the affect message is not necessarily reflected in divergence of polarity scores. We conducted an error analysis on instances where the SAM adjustment scores were not able to capture the MT's failure to transfer the correct  sentiment due to different linguistic phenomena. The first phenomenon we identified is related to structure shifting. For example, the sentiment distance between the source tweet \textit{`I was saddened by him'}  and its mistranslation \textit{`I made him sad'} is very small despite the flipping of sentiment direction in the translation. The three metrics with and without the SAM adjustment failed to penalise this type of distortion in the QE dataset. Second, some words in the lexicon did not have a score representative of their sentiment weight. For example, most prepositions in the SentiWord lexicon are neutral, yet by checking the data, it was found out that a mistranslation of a preposition can distort the affect message. For example, the source tweet \textit{`What is the benefit \textbf{of} me in this world'} was mistranslated by the MT system as \textit{`What is the benefit \textbf{for} me in this world'}; the wrong preposition causes the translation to lose the sad tone in the source tweet. Again, similar instances were not adequately measured with the three metrics with and without the SAM adjustment. Third, some nuanced sentiment-carrying words specific for the informal style of tweets caused a mistranslation of sentiment which was not captured by our approach. For instance, a one-word tweet referring to a political figure  as \textit{`Prick'} was mistranslated as \textit{`Sting'}. The source  is a slang word used to refer to a mean, contemptible man. The translation wrongly received a high score  since both `prick' and `sting' had very similar negative sentiment weights. The translation, however, fails to reflect the aggressive sentiment in the source tweet.  The BLEU metric, on the other hand,  succeeded in giving a penalty to similar short mistranslated tweets without the SAM adjustment.

One other limitation to the current approach for assessing the translation of sentiment is that it relies on an English sentiment lexicon. The applicability of this approach to other languages  depends on the availability of a similar high-precision and high-coverage sentiment lexicon. We have overcome this problem by using the English backtranslations of the Arabic and Spanish tweets in the QE dataset. There are, however, multilingual translations of English sentiment lexica that are commonly used  in the sentiment-analysis of non-English text. It remains to be tested how far a translated sentiment lexicon is capable of measuring sentiment transfer among different language pairs by the SAM scoring approach. 

\section{Conclusion}
\label{sec:concl}

 The most frequent scenario in which an MT system is used to translate sentiment-oriented text is for the translation of online UGC such as tweets, reviews or social media posts. The users of the MT online tool take these translations at face-value as there is no human intervention for accuracy checking. It is important, therefore, to ensure the reliability of the MT system to accurately transfer the author's affect message before it is used as an online tool. The reliability of an MT system with such big data is commonly measured by automatic metrics. We have shown that conventionally accepted metrics may not always be an optimum solution for assessing the translation of sentiment-oriented UGC. We presented an empirical approach to quantify the notion of sentiment transfer by an MT system and, more concretely, to modify automatic metrics such that its MT ranking comes closer to a human judgement of a poor or good translation of sentiment. Despite limitations to our approach, the SAM adjustment serves as a proxy to the complicated task of manually evaluating the translation of sentiment across different languages.




\section*{Acknowledgements}
This research work has been supported in part by the TranSent project at the Centre for Translation Studies (CTS) of the University of Surrey.

\bibliographystyle{acl_natbib}
\bibliography{anthology,ranlp2021}

\newpage
\begin{appendices}

\twocolumn[\section{Examples for Mistranslated Tweets Measured with and without SAM adjustment} \label{appendixa}]

\begin{figure}[h]

\centering
\includegraphics[scale=.54,trim={0cm .1cm 0cm 0cm},clip]{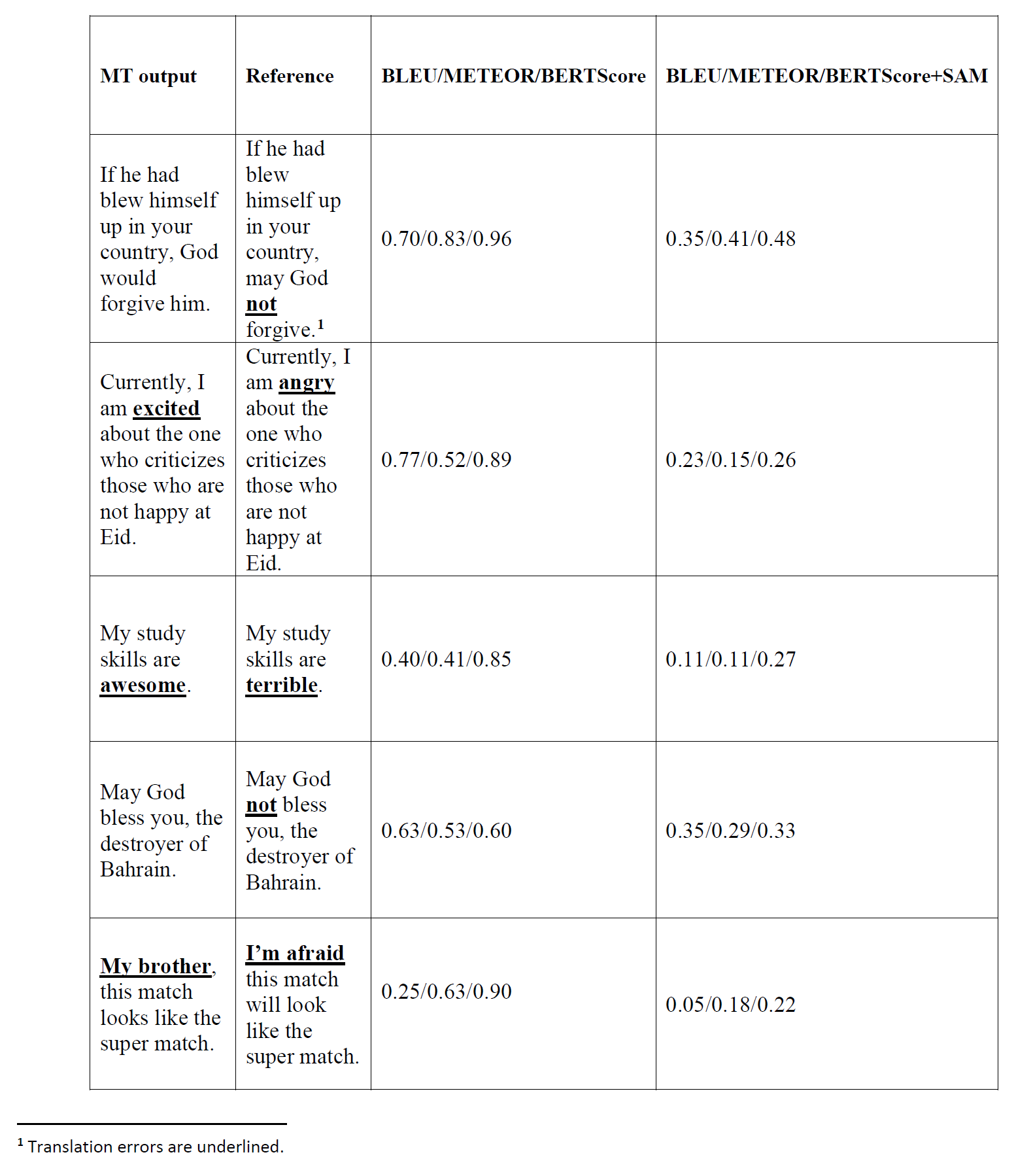}
\label{appendixpic}
\end{figure}
\end{appendices}

\end{document}